\documentclass{iopjournal}
\usepackage{amsmath}
\usepackage{booktabs}
\usepackage{subcaption}
\usepackage{hyperref, cite}
\usepackage[columnwise]{lineno}
\usepackage{doi}

\usepackage[numbers,sort&compress]{natbib}

\usepackage{enumitem}
\usepackage{amsfonts}
\usepackage{bm}
\usepackage{braket} 
\usepackage{lmodern}
\usepackage{multirow} 

\usepackage[dvipsnames]{xcolor}
\usepackage[normalem]{ulem}
\usepackage[columnwise]{lineno}

\usepackage{siunitx}
\begin{document}

\title{From Simulation to Experiment: Benchmarking Machine-Learning Models for Quantum-Dot State Recognition}

\author{
Valeria Díaz Moreno$^{1,*}$\orcid{0009-0001-2941-8019}
Ryan P. Khalili$^{1,*}$\orcid{0009-0000-6199-4713}, 
Daniel Schug$^{2,4,5}$\orcid{0009-0001-3758-501X}, 
Patrick J. Walsh$^{1, 3}$\orcid{0009-0004-6683-5817},\\
Donovan L. Buterakos$^{4,5}$\orcid{0000-0002-2286-3208},
Justyna P. Zwolak$^{4,5,6,\dagger}$\orcid{0000-0002-2286-3208}
}

\affil{$^1$Department of Physics, University of Wisconsin-Madison, Madison, Wisconsin 53706, USA}

\affil{$^2$Department of Computer Science, University of Maryland, College Park, MD 20742, USA}

\affil{$^3$Department of Applied Physics, Stanford University, Stanford, California 94305, USA}

\affil{$^4$National Institute of Standards and Technology, Gaithersburg, MD 20899, USA}

\affil{$^5$Joint Center for Quantum Information and Computer Science, University of Maryland, College Park, Maryland 20742, USA}

\affil{$^6$Department of Physics, University of Maryland, College Park, Maryland 20742, USA}

\affil{$^*$These authors contributed equally to this work.}

\affil{$^\dagger$Author to whom any correspondence should be addressed.}

\email{jpzwolak@nist.gov}

\begin{abstract}
Machine-learning models for quantum-dot state recognition are commonly trained using synthetic data, where large labeled datasets can be generated efficiently.
However, strong performance on simulated data does not guarantee equally strong performance in experiments.
Here, we benchmark five neural network architectures for quantum-dot state recognition using a shared synthetic training dataset and independent synthetic and experimental test sets.
We compare convolutional and fully connected neural networks, a mixture density network, a vision transformer, and a U-Net across multiple training set sizes and two input normalization schemes.
We find that model performance and rankings are closely preserved when moving from held-out synthetic patches to previously unseen simulated devices, but change substantially when the same models are evaluated on experimental data.
In particular, the architecture achieving the highest synthetic performance is not the one that performs best experimentally.
Input normalization further affects training behavior and, for some architectures, synthetic-to-experimental transfer.
These results show that synthetic benchmark performance alone is insufficient to select models for experimental deployment and highlight the need to jointly evaluate architecture, preprocessing, data efficiency, model complexity, and experimental transfer when developing machine-learning tools for automated quantum-dot characterization and control.
\end{abstract}

\section{Introduction}
\label{sec:intro}
Semiconductor quantum-dot (QD) spin qubits are a promising platform for quantum computing, combining long-lived spin degrees of freedom with electrostatic control in scalable device architectures~\cite{Hanson07-SQD, Kloeffel13-PQQ, Burkard21-SSQ, Vandersypen19-QCS}.
Operating these devices, however, requires coordinated control of many gate voltages to form the desired QD configuration and establish appropriate charge occupation and tunnel couplings.
Even tuning a double-QD (DQD) system is a nontrivial task because several electrostatically coupled gates must be adjusted simultaneously.
As QD arrays grow, device variability, gate cross-talk, charge rearrangements, and measurement overhead make manual and scripted tuning increasingly impractical, motivating the development of autonomous tuning approaches based on machine learning (ML)~\cite{Zwolak21-AAQ}.

\textit{Charge stability diagrams} (CSDs)~\cite{Wiel02-DQD}, two-dimensional (2D) measurements acquired while sweeping pairs of gate voltages, are central to many stages of QD tuning.
They are used to identify QD configurations~\cite{Zwolak20-AQD}, establish desired charge states~\cite{Durrer19-ATQ}, characterize device controllability~\cite{Ziegler22-TAR, Rao24-MAViS, Oakes24-AVE}, compensate for drift during operation~\cite{Rao25-TFS}, and detect unintended QD formation~\cite{Ziegler23-AEC}.
Transition patterns in CSDs encode information about QD formation, charge configuration, cross-capacitance, and interdot coupling, making automated interpretation of these measurements an important component of scalable device control.

A persistent challenge for ML-based CSD analysis is the availability of sufficiently large, labeled experimental datasets.
Although CSDs can be acquired routinely, expert annotation is time-consuming and can become ambiguous near boundaries between operating regimes.
Consequently, synthetic data have played an important role in the development of supervised ML methods for QD tuning, providing large datasets with automatically generated ground-truth labels and controllable models of device and measurement variability~\cite{Zwolak21-AAQ, qf-data}.
This approach has motivated both the release of publicly available QD datasets and a growing collection of open-source device simulators~\cite{Gualtieri25-QDsim, vanStraaten24-QAr, Krzywda25-QDa, Buterakos25-QDF}.
However, the use of synthetic training data raises a fundamental model-selection question: \emph{does performance on simulated validation or test data reliably identify the model that will perform best on experimental measurements?}

Previous work has begun to examine this question.
Darulov{\'a} \emph{et al.} compared ML models ranging from logistic regression to convolutional neural networks (CNNs) using synthetic, experimental, and mixed datasets for binary single- versus double-QD state recognition~\cite{Darulova20-EDM}.
They found that CNNs achieved the highest performance and that combining synthetic and experimental training data improved accuracy relative to experimental-only training.
They also demonstrated that performance depends on how faithfully simulated noise reproduces experimental conditions, highlighting the importance of the simulation-to-experiment gap.
Since then, increasingly diverse architectures---including deeper CNN models, encoder-decoder networks, and transformer-based approaches---have been introduced for QD image analysis.
It remains unclear, however, whether increasing model complexity improves experimental transfer, how model rankings depend on input preprocessing, and how these trade-offs evolve with the amount of available synthetic training data.

Here, we systematically benchmark five neural-network architectures for QD state recognition using a common synthetic training dataset and evaluation protocol.
We compare a multilayer perceptron (MLP), a convolutional neural network (CNN), a mixture density network (MDN), a vision transformer (ViT), and a U-Net across multiple synthetic data budgets and two input-normalization schemes.
All models are evaluated on three test sets that probe different generalization regimes: held-out synthetic patches drawn from the training-device population, synthetic data generated from previously unseen devices, and experimental CSDs.
This design allows us to distinguish performance on the synthetic benchmark from generalization across simulated devices and, ultimately, transfer to experiment.
In addition, we compare model complexity and training behavior and provide an open-source framework for reproducible benchmarking.

We find that model performance and rankings are closely preserved between the two synthetic evaluations, but change substantially when the same models are evaluated on experimental data.
In particular, testing on previously unseen simulated devices does not provide a better proxy for experimental model selection than the standard held-out synthetic benchmark.
These results show that generalization within the simulation domain and transfer to experimental measurements are distinct considerations when selecting ML models for QD state recognition.

The paper is organized as follows.
Section~\ref{sec:background} describes the QD state-recognition task, datasets, model architectures, and benchmarking protocol.
Section~\ref{sec:results} presents the comparison of synthetic and experimental model performance, data efficiency, training behavior, and model size.
We conclude in section~\ref{sec:conclusion} with implications for model selection in automated QD characterization and control.

\section{Background and methods}
\label{sec:background}
In this section, we define the quantum-dot state-recognition task and describe the datasets, model architectures, and common training and evaluation protocol used for benchmarking. 
We compare five neural-network architectures: MLP, CNN, MDN, ViT, and U-Net. 
All models receive the same \numproduct{30 x 30} input charge-stability-diagram (CSD) patch and produce five logits, which are converted with softmax to obtain the five-component fractional state vector, enabling direct comparison across architectures, training-set sizes, and normalization schemes. 
We conclude the section by describing the open-source benchmarking framework released with this work.

\subsection{Quantum-dot state-recognition task} 
\label{ssec:physics} 
\begin{figure}[t]
  \includegraphics[width=\linewidth]{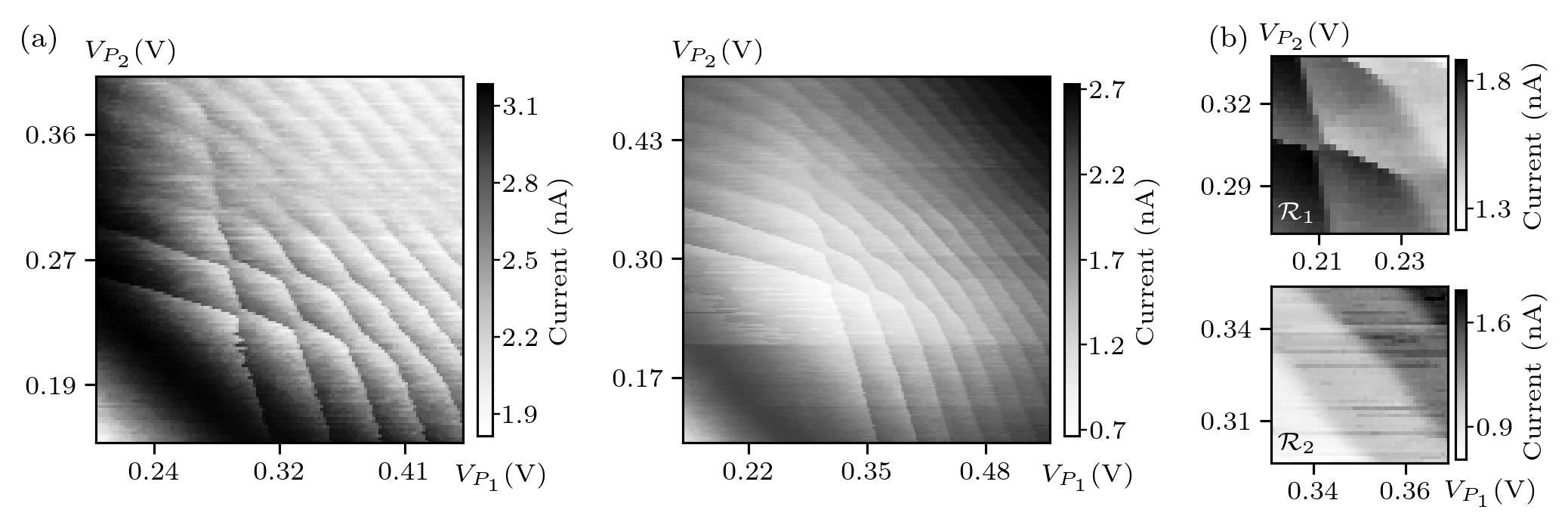}
  \caption{
  Examples of (a) two large and (b) two small experimentally acquired charge stability diagrams included in the \textit{QFlow~2.0: Quantum dot data for machine learning} dataset~\cite{qf-data}.
  The manually assigned fractional state vector labels for the small CSDs are ${\bf{p}}({\mathcal R}_1)=(0, 0, 0.1, 0, 0.9)$ and ${\bf{p}}({\mathcal R}_2)=(0, 0, 1, 0, 0)$.
  }
  \label{fig:data_exp}
\end{figure}

Semiconductor quantum dots (QDs) are formed by electrostatically confining charge carriers using lithographically defined gates~\cite{Vandersypen19-QCS, Burkard21-SSQ}. 
In a typical double-QD (DQD) device, so-called \textit{plunger gates} primarily control the electrochemical potentials of the individual dots, while \textit{barrier gates} control tunnel coupling to the reservoirs and between neighboring dots~\cite{Wiel02-DQD}. 
A central diagnostic for tuning these devices is the charge stability diagram (CSD), obtained by sweeping two gate voltages while monitoring the response of a proximal charge sensor~\cite{Simmons07-QCS}. 
Changes in charge occupation produce transition lines in the sensor response whose number, orientation, and connectivity provide information about QD formation and configuration. 
Representative experimental CSDs are shown in figure~\ref{fig:data_exp}(a). 

For the DQD state-recognition problem considered here, we distinguish five configurations commonly encountered during device tuning: no dot (ND), single left dot (SL), single central dot (SC), single right dot (SR), and double dot (DD). 
These configurations produce characteristic local patterns in CSDs.
Single-dot states are primarily distinguished by approximately parallel transition lines with different orientations, whereas the DD regime contains intersecting families of transitions that form a cell-like pattern. 
An example synthetic CSD and its corresponding state map are shown in figures~\ref{fig:data_sample}(a) and \ref{fig:data_sample}(b), respectively.
The ML task considered in this work is to infer the local QD configuration from a small region of a CSD.
Because such a region may span boundaries between different configurations, the target is represented by a five-component fractional state vector rather than a single state label; its construction is described in section~\ref{ssec:data}.

Reliable identification of these states is an important component of automated QD tuning. 
The gate-voltage ranges associated with a given configuration vary across devices and cooldowns because of device variability, electrostatic cross-coupling, and changes in the local charge environment. 
Previous work has demonstrated that ML models trained exclusively on simulated CSDs incorporating realistic device and measurement variability can successfully recognize states in experimental data~\cite{Zwolak21-AAQ}. 
Here, we use this synthetic-to-experimental setting to systematically compare how different ML model architectures perform under a common training and evaluation framework.

\subsection{Synthetic and experimental datasets}
\label{ssec:data}
The synthetic data used in this work were generated with \texttt{QDFlow}, a Python package for physics-based simulations of QD devices~\cite{Buterakos25-QDF, Buterakos25-QDFcode}. 
\texttt{QDFlow} uses a modified Thomas-Fermi approximation to model a quasi-one-dimensional (1D) nanowire controlled by five electrostatic gates, whose voltages determine QD formation, charge occupancy, and conductance.
The dataset consists of \num{1000} simulated devices, each represented by a noiseless, static CSD whose transition geometry is determined by the corresponding device parameters.
For each device, CSDs are generated as they would be measured by sensors at three locations: the left, center, and right sides of the device.
At each sensor location, \num{10} random noise realizations are applied to the same underlying noiseless CSD, representing repeated measurements of a fixed device configuration.
Each noise realization includes white, pink, and random-telegraph noise with varying amplitudes, together with measurement effects such as sensor-gate coupling, Gaussian blurring, nonlinear Coulomb-peak response, unintended dot transitions, and latching.
The parameters governing these effects were chosen to qualitatively reproduce features observed in experimental QD measurements.
Each CSD is stored as a \numproduct{251 x 251} array, along with a pixel-wise state map that identifies the QD configuration, the corresponding plunger-gate values $(V_{P_1}, V_{P_2})$, and metadata describing the simulated device and noise parameters.

\begin{figure}[t]
  \includegraphics[width=\linewidth]{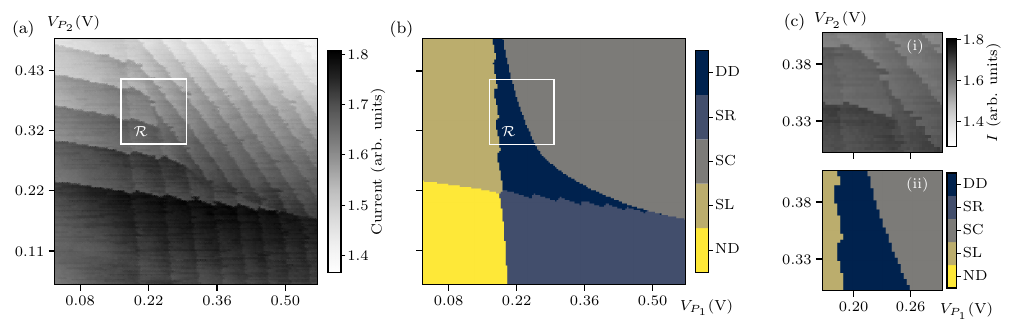}
  \caption{
  (a) Example synthetic charge stability diagram and (b) the corresponding state map generated using \texttt{QDFlow}~\cite{Buterakos25-QDF, Buterakos25-QDFcode}.
  The state-map labels correspond to the five states possible for a double-QD device: no dot (ND), single left dot (SL), single central dot (SC), single right dot (SR), and double dot (DD).
  (c-i) Example patch ${\mathcal R}$ sampled from the CSD, highlighted in panel (a) with a white rectangle.
  (c-ii) Corresponding state-map patch, highlighted in panel (b) with a white rectangle.
  The fractional state vector assigned to this patch is ${\bf{p}}({\mathcal R})=(0, 0.1, 0.5, 0, 0.4)$.
  }
  \label{fig:data_sample}
\end{figure}

To construct the ML dataset, five \numproduct{30 x 30}-pixel patches are randomly sampled from each synthetic CSD, yielding \num{150000} samples.
The patch size was selected to remain compatible with the experimentally acquired small CSDs included in the \textit{QFlow~2.0} dataset, which range from approximately \numproduct{30 x 30} to \numproduct{60 x 60} pixels with resolutions of approximately \qtyrange{1}{2}{\milli\volt} per pixel~\cite{qf-data}.
Each synthetic sample contains the CSD patch, the corresponding state-map patch, local voltage coordinates, the patch's position within the parent CSD, the parent CSD identifier, and the associated noise identifier.
An example CSD patch and its corresponding state-map patch are shown in figures~\ref{fig:data_sample}(c-i) and \ref{fig:data_sample}(c-ii), respectively.

Because a small CSD region can span boundaries between operating regimes, a single patch may contain more than one QD state.
We therefore represent the ground truth using a five-component fractional state vector,
\begin{equation}
\label{eq:prob_vec}
\mathbf{p}({\mathcal R}) = \left(p_{\rm ND}, p_{\rm SL}, p_{\rm SC}, p_{\rm SR}, p_{\rm DD}\right),
\end{equation}
where $p_{\rm X}$ is the fraction of pixels in the corresponding state-map patch assigned to state ${\rm X}\in\{{\rm ND},{\rm SL},{\rm SC},{\rm SR},{\rm DD}\}$, such that $\sum_{\rm X}p_{\rm X}=1$.
For example, the patch shown in figure~\ref{fig:data_sample}(c) contains both SR and DD regions and is assigned
$\bm{p}({\mathcal R})=(0, 0.1, 0.5, 0, 0.4)$.
Thus, the fractional labels describe the spatial composition of the sampled voltage region rather than uncertainty in its ground-truth annotation~\cite{Kalantre17-MLD}.

From the complete set of \num{150000} synthetic patches, we randomly reserve \num{10500} as a held-out synthetic test set.
Of the remaining \num{139500} samples, \qty{90}{\percent} (\num{125550}) are used as the available training set and \qty{10}{\percent} (\num{13950}) are reserved as a common validation set.
To study data efficiency, models are trained using five data budgets corresponding to \qty{20}{\percent}, \qty{40}{\percent}, \qty{60}{\percent}, \qty{80}{\percent}, and \qty{100}{\percent} of the available training set, or \num{25110}, \num{50220}, \num{75330}, \num{100440}, and \num{125550} samples, respectively.
The training and partitioning procedure used for each data budget is described in section~\ref{ssec:training}.

The split into training, validation, and held-out synthetic test sets is performed at the patch level after sampling.
Consequently, distinct training and test patches may originate from the same simulated device or parent CSD.
The held-out synthetic test set therefore measures generalization to previously unseen local CSD regions drawn from the same simulated-device population represented during model development.

To separately test generalization to unseen simulated devices, we generate an additional \num{50} devices using the same simulation procedure and parameter distributions as for the development dataset.
Sampling these devices in the same manner yields a device-held-out synthetic test set containing \num{7500} patches.
None of these devices is represented in the training, validation, or held-out synthetic test sets, and the associated noise realizations are sampled independently.
Because the device and noise parameters are drawn from the same underlying distributions used for the development data, this test set probes generalization to new simulated devices within the same simulation domain rather than transfer to a different data-generating distribution.

Finally, synthetic-to-experimental transfer is evaluated using \num{756} manually labeled small experimental CSDs included in \textit{QFlow~2.0}~\cite{qf-data}.
The experimental data were acquired using two distinct QD devices, with one measured over two thermal cycles and the other over a single thermal cycle.
These data are not used for model training, validation, or model selection and provide a common experimental evaluation set for all architectures and training conditions.
Representative experimental CSDs and manually assigned fractional state vectors are shown in figure~\ref{fig:data_exp}(b).

Because the three evaluation datasets are sampled independently, their marginal distributions over the five dominant QD states are not identical. 
To reduce the influence of these differences on cross-dataset performance comparisons, the two synthetic test sets are evaluated using importance weights that match their dominant-state distributions to that of the experimental test set. 
The underlying test sets and their sample sizes are unchanged; the weighting is applied only when computing evaluation metrics, as described in section~\ref{ssec:evaluation}.

\subsection{Input representation and normalization}
\label{ssec:preprocessing}
We use the one-dimensional (1D) derivative of each CSD along the measurement direction, rather than the raw sensor signal, as model input.
The derivative emphasizes the charge-transition lines that distinguish the different QD configurations while suppressing slowly varying background contributions, making the noise floor less pronounced relative to the transition features.
All resulting inputs retain the original \numproduct{30 x 30}-pixel resolution and are supplied to the models as single-channel images.

To investigate the effect of input normalization, each model is trained using two normalization schemes while keeping the training subsets and data partitions fixed: z-score standardization and min-max scaling.
These transformations provide two different ways of rescaling the signed derivative signal: z-score standardization centers each patch and scales it by its standard deviation, whereas min-max scaling maps the local extrema to a fixed interval. 
Comparing the two, therefore, allows us to assess how strongly model performance depends on the representation of signal amplitude within individual CSD patches.

For an input patch $x$, z-score standardization is defined as
\begin{equation}
    x_{\rm z}=\frac{x-\mu}{\sigma},
    \label{eq:zscore}
\end{equation}
where $\mu$ and $\sigma$ are the mean and standard deviation calculated over the individual CSD patch.
Min-max scaling is defined as
\begin{equation}
    x_{\rm mm}=\frac{x-x_{\min}}{x_{\max}-x_{\min}},
    \label{eq:minmax}
\end{equation}
where $x_{\min}$ and $x_{\max}$ are the minimum and maximum values within the individual CSD patch, respectively, such that the transformed values lie between \num{0} and \num{1}.
The same preprocessing procedure is applied to synthetic and experimental CSDs for a given training condition.
Each model architecture is therefore benchmarked using both normalization schemes.

\subsection{Model architectures} 
\label{ssec:models}
\begin{figure}[t]
  \includegraphics[width=\linewidth]{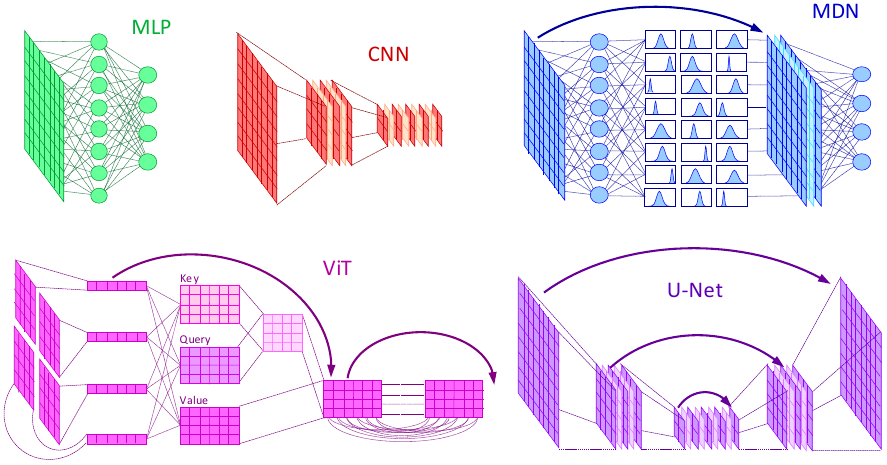}
  \caption{
  Schematic representations of the five model architectures considered in this work.
  Each architecture receives the same \numproduct{30 x 30} preprocessed CSD patch as input and produces five output logits corresponding to the ND, SL, SC, SR, and DD states.
  The schematics illustrate the principal architectural differences among the models; implementation details are provided in section~\ref{ssec:models}.
  }
  \label{fig:models}
\end{figure}

We benchmark five neural-network architectures that represent distinct approaches to extracting information from image-like inputs: an MLP, a CNN, an MDN, a ViT, and a U-Net.
The models differ substantially in how they represent spatial information, ranging from a fully connected MLP that operates on flattened inputs to convolutional, attention-based, and encoder-decoder architectures.
The architectures are illustrated schematically in figure~\ref{fig:models}.

For the MLP, CNN, ViT, and U-Net, we use canonical implementations of the corresponding architecture families, adapted only as needed to accommodate the common \numproduct{30 x 30} input and five-component output used in this work.
For the MDN, for which there is no comparably canonical image-classification architecture, we follow the published MDN-C1 classification formulation of Gugulothu \emph{et al.}~\cite{Gugulothu24-MDC}.
The resulting benchmark therefore compares practical representative model configurations rather than attempting to identify an independently optimized architecture for each model family.

All models receive a single-channel \numproduct{30 x 30} preprocessed CSD patch and produce five logits corresponding to the ND, SL, SC, SR, and DD states.
Finally, softmax is used to transform the logits into a five-component fractional state vector, which is returned during inference or used to compute the KL divergence loss during training.

\paragraph{Multilayer perceptron.}
The MLP provides a simple fully connected baseline that does not explicitly encode spatial locality.
Each \numproduct{30 x 30} CSD patch is flattened into a \num{900}-dimensional input vector and passed through two hidden layers with \num{256} and \num{128} units, respectively.
Each hidden layer uses Gaussian error linear unit (GELU) activation and batch normalization, with dropout applied at a probability of \num{0.3} for regularization.
The final fully connected layer produces the five output logits.
MLPs have previously been applied to charge-state recognition and autotuning of QD devices~\cite{Zwolak21-RBI, Czischek21-MNA}.

\paragraph{Convolutional neural network.}
CNNs use learned local filters to extract spatial features from image-like inputs, making them a natural choice for identifying the transition-line patterns present in CSDs~\cite{Krizhevsky12-CNN}.
They have previously been applied to QD state recognition and data-quality assessment in automated tuning workflows~\cite{Zwolak18-QLD, Ziegler22-TRA, Yon24-ECA}, as well as to device-usability assessment through detection of fabrication defects in scanning electron micrographs~\cite{Mei20-OQF}.
The CNN used here is based on previous demonstrations of CNN-based QD autotuning~\cite{Ziegler22-TRA}.

The network contains four convolutional layers with \num{32}, \num{64}, \num{128}, and \num{128} output channels, respectively.
Each of the first three convolutional layers is followed by max pooling, which reduces the spatial resolution by a factor of \num{2}, along with batch normalization and ReLU activation.
Adaptive average pooling is applied after the final convolutional layer, followed by a fully connected layer that produces the five output logits.

\paragraph{Mixture density network.}
MDNs combine a feed-forward neural network with a parametric mixture representation~\cite{Bishop94-MDN, Makansi19-MDN}.
Unlike the other model families considered here, there is no standard MDN architecture for image classification.
We therefore follow the MDN-C1 classification formulation introduced in Ref.~\cite{Gugulothu24-MDC}, adapting its input and output dimensions to the present state-recognition task.
In this formulation, the learned mixture distributions provide an intermediate representation of the input, and values derived from their cumulative distribution functions are subsequently used for classification.

Each \numproduct{30 x 30} CSD patch is flattened into a \num{900}-dimensional input vector and passed through a \num{512}-unit hidden layer followed by layer normalization.
The network then predicts the parameters of a three-component Gaussian mixture for each of the \num{900} input dimensions.
For each dimension, the corresponding Gaussian cumulative-distribution-function values are weighted by the learned mixture coefficients, and the resulting values are concatenated to form a \numproduct{900 x 3}\,=\,\num{2700}-dimensional representation.
This representation is passed through a \num{256}-unit fully connected layer followed by a five-unit output layer.
GELU activation and dropout with probability \num{0.2} are applied to the hidden fully connected layers.
Positive standard deviations are enforced using a softplus transformation with an offset of $\epsilon=\num{1e-5}$ for numerical stability.

\paragraph{Vision transformer.}
ViTs represent images as sequences of local patches and use self-attention to model relationships among them~\cite{Vaswani17-AAN, Dosovitskiy21-TIR}.
For the \numproduct{30 x 30} CSD inputs used here, each image is divided into non-overlapping \numproduct{5 x 5}-pixel patches, producing \num{36} image tokens.
Each token is linearly embedded in a \num{128}-dimensional space and augmented with positional embeddings, together with a learnable class token.
The network contains six transformer encoder blocks.
Each block uses four-head self-attention followed by a feed-forward network with dimensions \num{128}, \num{512}, and \num{128}, with residual connections around both components.
After the final transformer block, the class-token representation is passed through a classification MLP with dimensions \num{256}, \num{128}, and \num{5}.
Dropout is applied with probability \num{0.1} within the transformer and classification blocks.
This architecture tests whether attention-based interactions among spatially separated regions of a CSD provide an advantage over architectures with a predominantly local inductive bias.

\paragraph{U-Net.}
U-Net combines a convolutional encoder-decoder architecture with skip connections that propagate high-resolution features between corresponding encoder and decoder stages~\cite{Ronneberger15-UNB, Tai25-MEU}.
This structure has previously been applied to several QD image-analysis tasks, including detection of unintended QDs, device virtualization, transition-line detection, and compensation for drift~\cite{Ziegler23-AEC, Rao24-MAViS, Hader25-CTD, Rao25-TFS}.

Our implementation contains two downsampling stages using \numproduct{2 x 2} max pooling and two corresponding upsampling stages using transposed convolutions.
The encoder increases the number of feature channels from \num{64} to \num{128}, followed by a \num{256}-channel bottleneck.
The decoder then reduces the number of channels from \num{256} to \num{128} and finally to \num{64}, with skip connections linking corresponding encoder and decoder stages.
Each encoder, decoder, and bottleneck block contains two \numproduct{3 x 3} convolutional layers with ReLU activations and spatial dropout with probability \num{0.2}.
A final convolutional layer reduces the representation to five feature channels corresponding to the five QD states.
Because the present task requires a patch-level fractional state prediction rather than pixel-wise segmentation, adaptive average pooling is applied to these five feature maps to obtain the final five-component output.

\subsection{Training protocol and data-efficiency study}
\label{ssec:training}
To assess data efficiency, we train each architecture using five data budgets corresponding to \qty{20}{\percent}, \qty{40}{\percent}, \qty{60}{\percent}, \qty{80}{\percent}, and \qty{100}{\percent} of the \num{125550}-sample synthetic training set.
For each data budget, a single training subset is selected and held fixed across all model architectures and normalization schemes.
The training subset is divided into \num{10} stratified partitions, and each of the \num{10} models is trained on nine of the \num{10} partitions, with a different partition held out for each model.
The independently set-aside \num{13950}-sample validation set is shared across all models, data budgets, architectures, and normalization schemes.

Stratification is performed according to the dominant state of each sample, defined as the state corresponding to the largest component of the fractional target vector, $\arg\max_j p_j$.
For example, a target vector $(0,0.7,0,0,0.3)$ is assigned to the SL stratum.
Random sampling and partition assignments are seeded for reproducibility, and the same training subsets and partition assignments are used across model architectures and normalization schemes.
The common validation set is used to monitor training, apply the early-stopping criterion, and select the best checkpoint for each model.
The held-out synthetic, device-held-out synthetic, and experimental test sets described in section~\ref{ssec:data} are not used during training, validation, or model selection.

All models are optimized using AdamW~\cite{Loshchilov17-DWR}.
The learning rate and weight decay are adjusted empirically for each architecture based on training and validation behavior, with the resulting model-specific values given in table~\ref{tab:hyperpars}.
A cosine-annealing learning-rate scheduler is used for the MLP, MDN, ViT, and U-Net, whereas the CNN is trained using a fixed learning rate.
The training objective is the Kullback-Leibler (KL) divergence between the ground-truth and predicted fractional state vectors.
Because both vectors are normalized and non-negative, the KL divergence provides a natural distributional objective for learning the fractional composition of the five QD states.
For a minibatch containing $B$ samples, the loss is
\begin{equation}
    \mathcal{L}_{\rm KL} = \frac{1}{B}\sum_{i=1}^{B}\sum_{j=1}^{5}p_{ij}
    \log\left(\frac{p_{ij}}{\hat{p}_{ij}}\right),
    \label{eq:kl_loss}
\end{equation}
where $\bm{p}_i$ and $\hat{\bm{p}}_i$ denote the ground-truth and predicted fractional state vectors, respectively, and the predicted vector is obtained by applying softmax to the five model logits.
Terms with $p_{ij}=0$ are taken to contribute zero to the sum.

Training is allowed to proceed for up to \num{150} epochs and is terminated early based on the validation KL-divergence loss.
Early stopping is triggered when the validation loss fails to improve by at least $10^{-4}$ for \num{10} consecutive epochs.
All training runs considered here terminate through the early-stopping criterion before reaching the maximum number of epochs.
The checkpoint with the lowest validation KL divergence loss is retained for subsequent evaluation.
Thus, model training and checkpoint selection are governed entirely by the KL-divergence objective rather than by the MSE metric used for the primary performance comparison.

\begin{table}[t]
\caption{AdamW optimizer hyperparameters and learning-rate schedules used for the five model architectures.}
\centering
\begin{tabular}{lccccc}
\hline
Hyperparameter & MLP & CNN & MDN & ViT & U-Net \\
\hline
Learning rate $\eta$
    & \num{1e-3} & \num{5e-4} & \num{1e-3} & \num{1e-4} & \num{5e-4} \\
Weight decay
    & \num{1e-4} & \num{2e-4} & \num{1e-4} & \num{3e-4} & \num{1e-4} \\
Dropout
    & \num{0.3} & \num{0} & \num{0.2} & \num{0.1} & \num{0.2} \\
Activation
    & GELU & ReLU & GELU & GELU & ReLU \\
LR schedule
    & Cosine & Fixed & Cosine & Cosine & Cosine \\
\hline
\end{tabular}
\label{tab:hyperpars}
\end{table}

\subsection{Evaluation protocol and performance metrics}
\label{ssec:evaluation}
Model performance is evaluated on three datasets that are not used for training or validation: the held-out synthetic test set, the device-held-out synthetic test set, and the experimental test set described in section~\ref{ssec:data}.
The first measures generalization to previously unseen CSD patches drawn from the same simulated-device population as the training data, the second measures generalization to entirely unseen simulated devices and noise realizations, and the third measures synthetic-to-experimental transfer.
For each architecture, normalization scheme, and data budget, each of the \num{10} resulting models is evaluated separately on all three test datasets.

Because the relative prevalence of the five QD states varies across the evaluation datasets, we apply importance weighting when evaluating the two synthetic test sets. 
Each sample is assigned a dominant state,
\begin{equation}
    s_i = \arg\max_j p_{ij},
\end{equation}
consistent with the stratification used during model training. 
For a synthetic test set, the weight assigned to a sample with dominant state $s_i$ is
\begin{equation}
    w_i = \frac{P_{\rm exp}(s_i)}{P_{\rm sim}(s_i)},
\end{equation}
where $P_{\rm exp}(s)$ and $P_{\rm sim}(s)$ denote the empirical frequencies of dominant state $s$ in the experimental and synthetic test sets, respectively. 
This weighting makes the marginal distribution of the five dominant states in each synthetic evaluation equivalent to that of the experimental test set while retaining all available synthetic samples.

Since the target labels describe the fractional composition of the five QD states within a CSD patch, model performance is quantified using the observation-wise mean-squared error (MSE) between the predicted and ground-truth state vectors.
Before evaluation, the five model logits are transformed using softmax to obtain the predicted fractional state vector $\hat{\bf p}_i$.
For a model evaluated on a dataset containing $N$ samples, we define weighted MSE as
\begin{equation}
    {\rm MSE} = \sum_{i=1}^{N}w_i \left[\sum_{j=1}^{5} 
    \left(\hat{p}_{ij}-p_{ij}\right)^2\right] \times \left(\sum_{i=1}^{N} w_i\right)^{-1}
    \label{eq:mse}
\end{equation}
The experimental test set is evaluated without reweighting, corresponding to $w_i=1$ for all samples.
For ease of interpretation, we report the corresponding MSE score, $S_{\rm MSE}=1-{\rm MSE}$, such that larger values indicate better agreement between the predicted and ground-truth state compositions.
The MSE score is calculated independently for each of the \num{10} trained models.

To provide a reference for the MSE score scale, we additionally consider image-independent baseline predictions.
A uniform baseline assigns equal weight to all five states,
\begin{equation}
    \hat{\bf p}_{\rm uniform} = (0.2,0.2,0.2,0.2,0.2),
\end{equation}
for every input.
We also consider an empirical baseline given by the mean state-composition vector of the corresponding synthetic training set,
\begin{equation}
    \hat{\bf p}_{\rm mean} =
    \frac{1}{N_{\rm train}} \sum_{i=1}^{N_{\rm train}}{\bf p}_i.
\end{equation}
The latter represents the optimal constant, image-independent prediction under the MSE objective for the training distributions.
Both baselines are evaluated on all three test datasets using the same MSE score as the learned models.

To quantify the correspondence in predictive performance across test datasets, we compare the mean $S_{\rm MSE}$ obtained for each matched combination of model architecture, normalization scheme, and training-data budget.
The five architectures, two normalization schemes, and five data budgets yield \num{50} matched conditions for each pair of evaluation datasets.
We report the Pearson correlation coefficient, $r$, to quantify the linear correspondence between condition-wise scores and the Spearman rank correlation coefficient, $\rho$, to quantify the extent to which their relative ordering is preserved.
We additionally report the mean signed difference in $S_{\rm MSE}$ between datasets to identify systematic shifts in overall performance.
These quantities are used descriptively to compare evaluation domains rather than as hypothesis tests.

\subsection{Training characteristics and model-selection analysis}
\label{ssec:compute}
In addition to predictive performance, we compare model complexity and training behavior across the five architectures.
Model complexity is characterized by the number of trainable parameters, calculated directly for each implemented architecture.
We further characterize training behavior by the number of epochs required before the early-stopping criterion described in section~\ref{ssec:training} is reached.
For each architecture, normalization scheme, and data budget, this quantity is recorded independently for each of the \num{10} trained models and summarized across them.

To compare predictive performance with model size, we construct Pareto frontiers using the mean MSE score, $S_{\rm MSE}$, and the number of trainable parameters.
Here, parameter count is used as a measure of model size rather than computational cost, which also depends on factors such as parameter reuse, operation count, memory access, and implementation details.
Because the architectures are evaluated using canonical or reference configurations rather than through architecture-specific width or depth optimization, these frontiers compare the particular model configurations considered here and do not represent optimized performance-size frontiers for the corresponding architecture families.

For a fixed training-data budget and normalization scheme, a model is considered Pareto optimal if no other architecture achieves an equal or higher $S_{\rm MSE}$ with an equal or smaller number of trainable parameters, with at least one of these inequalities being strict.
Models on the Pareto frontier, therefore, represent architectures for which improved predictive performance cannot be obtained without increasing model size.

To assess the robustness of Pareto-frontier membership to model-to-model variability, we additionally perform a bootstrap analysis using the \num{10} trained models for each condition.
For each bootstrap realization, the \num{10} training-partition indices are resampled with replacement, preserving the correspondence among architectures, and the mean score and corresponding Pareto frontier are recomputed.
The mean $S_{\rm MSE}$ for each architecture and the corresponding Pareto frontier are then recomputed.
We use \num{50000} bootstrap realizations for each dataset, normalization scheme, and training-data budget.
We report the bootstrap Pareto-membership probability, $P_{\rm F}$, defined as the fraction of bootstrap realizations in which a given architecture lies on the Pareto frontier.

To facilitate reproducibility and future comparison, we provide an open-source implementation of the complete benchmarking workflow~\cite{ML-suite}.
The framework provides a common training and evaluation pipeline for the MLP, CNN, MDN, ViT, and U-Net architectures considered here.
Experiment settings are specified through human-readable configuration files, including the model architecture, optimizer, learning-rate scheduler, normalization scheme, data budget, and early-stopping parameters.
Seeded sampling and partition assignments reproduce the data subsets and training partitions used in this study.

A common command-line interface controls data preparation, training, and evaluation, and records the configuration associated with each run.
The framework stores the best-performing model checkpoints, training and validation histories, per-model evaluation metrics, and machine-readable output files for subsequent analysis and aggregation.
Together, these features provide a unified and reproducible framework for extending the benchmark to additional model architectures, preprocessing strategies, and QD datasets.

\section{Results}
\label{sec:results}
We organize the benchmarking results around three questions: (i) how much synthetic training data is needed to achieve stable predictive performance, (ii) how well performance on synthetic data predicts performance on experimental data, and (iii) what model-selection trade-offs emerge when predictive performance is considered together with model size.
We first examine predictive performance as a function of training data budget and normalization scheme, thereby allowing us to compare the data efficiency of the five model architectures.
We then compare performance across the held-out synthetic, device-held-out synthetic, and experimental test sets to determine how reliably synthetic benchmarks predict experimental model performance.
Finally, we examine training behavior and model size, and consider these characteristics alongside predictive performance in a Pareto analysis.

\subsection{How much synthetic training data is needed?}
\label{ssec:data_dependence}
\begin{figure}[t]
  \includegraphics[width=\linewidth]{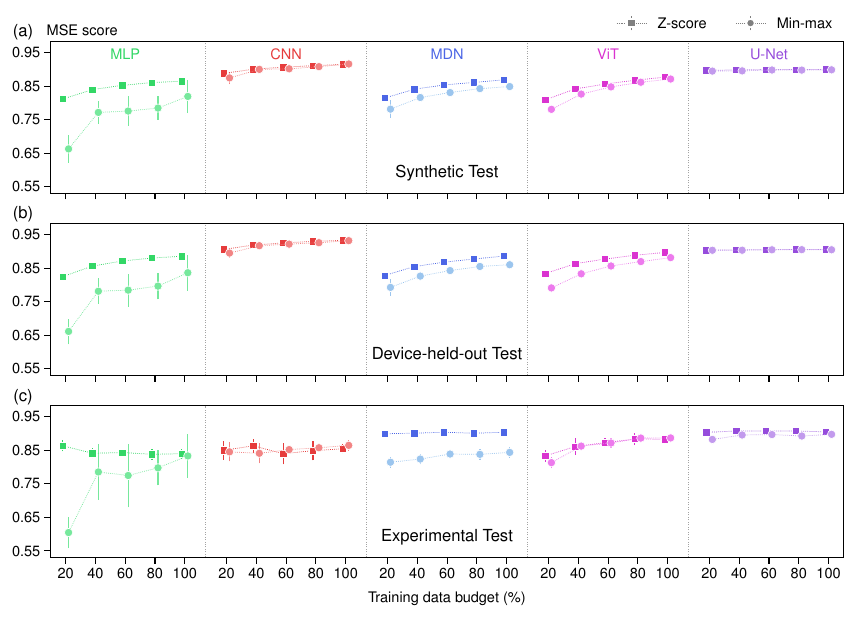}
  \caption{
  MSE score, $S_{\rm MSE}=1-\mathrm{MSE}$, as a function of synthetic training-data budget for models evaluated on
  (a) the held-out synthetic test set,
  (b) the device-held-out synthetic test set, and
  (c) the experimental test set.
  For panels (a) and (b), synthetic samples are importance-weighted such that the marginal distribution of dominant QD states matches that of the experimental test set; panel (c) shows the experimental data without reweighting. 
  Within each architecture, points correspond to training-data budgets of \qty{20}{\percent}, \qty{40}{\percent}, \qty{60}{\percent}, \qty{80}{\percent}, and \qty{100}{\percent}.
  Colors identify the model architecture, while marker styles distinguish z-score (squares) and min-max (circles) normalization.
  Points show the mean performance across the \num{10} trained models for each condition, and error bars indicate one standard deviation across the \num{10} model-level scores, providing a measure of model-to-model variability.
  }
  \label{fig:perf}
\end{figure}

We first examine how the amount of synthetic training data affects predictive performance and model-to-model consistency across architectures and normalization schemes.
Figure~\ref{fig:perf}(a) shows the MSE score, $S_{\rm MSE}=1-\mathrm{MSE}$, on the held-out synthetic test set for models trained using five data budgets ranging from \qty{20}{\percent} to \qty{100}{\percent} of the available synthetic training data.
For reference, the image-independent uniform and training-distribution baselines perform substantially below the learned models across all three evaluation datasets; the corresponding $S_{\rm MSE}$ values are summarized in table~\ref{tab:baselines}.
All trained model configurations substantially outperform the corresponding baselines, with even the lowest mean score exceeding the uniform baseline by approximately \num{0.33} and the training-data mean baseline by approximately \num{0.32}.

The effect of increasing the training-data budget varies noticeably across architectures.
The CNN achieves the highest synthetic test score across the full range of training-data budgets and improves modestly as additional training data are introduced.
In contrast, the U-Net reaches nearly its maximum performance with only \qty{20}{\percent} of the available training data and shows little further improvement as the data budget increases.
The MLP, MDN, and ViT exhibit a stronger dependence on training-set size, with their mean scores increasing progressively as larger synthetic datasets are used for training.
These differences show that the amount of training data required to reach stable performance is architecture dependent: some models effectively saturate at relatively small data budgets, whereas others continue to benefit from additional synthetic examples.

\begin{table}[t]
\caption{
MSE scores for the image-independent baseline predictors on the three evaluation datasets.
The training dataset mean was established based on the full 
Synthetic test-set scores use the dominant-state importance weighting described in the text; the experimental test set is evaluated without reweighting.
}
\centering
\begin{tabular}{lccc}
\hline
Baseline & Held-out synthetic & Device-held-out synthetic & Experimental \\
\hline
Uniform & \num{0.326} & \num{0.316} & \num{0.277} \\ 
Training dataset mean & \num{0.337} & \num{0.333} & \num{0.286} \\
\hline
\end{tabular}
\label{tab:baselines}
\end{table}

Performance on the device-held-out synthetic test set, shown in figure~\ref{fig:perf}(b), closely follows the same trends.
Across architectures, normalization schemes, and training-data budgets, withholding complete simulated devices produces only a small shift in absolute score and largely preserves the relative performance observed on the held-out synthetic patches.
Thus, the architecture-dependent data-efficiency trends are not specific to evaluating on new patches drawn from simulated devices represented in the training data, but persist when the models are evaluated on previously unseen simulated devices.

Normalization further affects both mean predictive performance and model-to-model variability.
The strongest dependence is observed for the MLP.
With z-score normalization, the MLP's mean score increases steadily with training-data budget, while variation among the \num{10} trained models remains comparatively small.
Under min-max normalization, the MLP achieves lower mean performance and substantially larger model-to-model variability, particularly at smaller and intermediate data budgets.
The MDN also performs better with z-score normalization, although its performance improves with increasing data budget under both normalization schemes.
By comparison, the CNN and U-Net are relatively insensitive to normalization, with similar mean scores and trends for the two preprocessing schemes.
The ViT likewise improves as the data budget increases under both normalization schemes.

Overall, these results show that data efficiency is not a universal property of the state-recognition task, but depends strongly on model architecture and input normalization.
Additional synthetic training data provide limited benefit once the U-Net reaches its early performance plateau, whereas the MLP, MDN, and ViT continue to improve over the range considered.
At the same time, the large model-to-model variability of the min-max-normalized MLP indicates that increasing the training-set size alone does not guarantee stable performance when the normalization scheme is poorly matched to the architecture.
The close agreement between the two synthetic evaluations further indicates that these conclusions remain largely unchanged when generalization is tested on entirely unseen simulated devices.

\subsection{Does synthetic performance predict experimental performance?}
\label{ssec:synthetic_to_experiment}
Performance on held-out synthetic data provides a natural basis for selecting a model architecture, normalization scheme, and training-data budget.
We next ask whether the relative performance observed on this benchmark persists when the models are evaluated on previously unseen simulated devices and, ultimately, on experimental data.
Figure~\ref{fig:perf} compares all trained models across these three evaluation sets.

We first consider the device-held-out synthetic test set; see figure~\ref{fig:perf}(b).
Across architectures, normalization schemes, and training-data budgets, performance closely tracks that obtained on the held-out synthetic patches.
The condition-wise scores on the two synthetic test sets are strongly correlated (Pearson $r=0.994$; Spearman $\rho=0.989$), with the device-held-out scores exhibiting only a small systematic increase of approximately \num{0.014} in $S_{\rm MSE}$ on average.
Thus, withholding complete simulated devices does not materially change the conclusions obtained from the standard synthetic benchmark.
The CNN remains the highest-performing architecture across the full range of training-data budgets, followed closely by the U-Net, and the architecture-dependent trends with training-data budget and normalization are largely preserved.

This correspondence changes substantially when the same models are evaluated on experimental CSDs; see figure~\ref{fig:perf}(c).
Although the held-out synthetic and experimental scores retain a moderate linear correspondence (Pearson $r=0.765$), their rank correspondence is substantially weaker (Spearman $\rho=0.506$).
Thus, conditions that perform well synthetically tend, in broad terms, to perform well experimentally, but their relative ordering is considerably less well preserved.
Using the device-held-out synthetic test set does not improve this correspondence: its correlation with experimental performance is similar in magnitude ($r=0.753$; $\rho=0.464$).
Evaluating on previously unseen simulated devices therefore does not, by itself, provide a better proxy for experimental model selection.
Because the synthetic scores are adjusted to match the experimental marginal dominant-state distribution, the weaker synthetic-to-experimental correspondence cannot be attributed solely to differences in the relative prevalence of the five dominant QD states.

The change is also evident in the relative performance of the individual architectures.
On both synthetic test sets, the CNN achieves the highest mean $S_{\rm MSE}$ across normalization schemes and training-data budgets, while the U-Net provides similarly strong performance with little dependence on training-set size.
On experimental data, however, the CNN no longer provides the strongest performance.
For z-score-normalized inputs, the MDN and U-Net achieve the highest experimental scores despite both ranking below the CNN on synthetic data.
With min-max normalization, the U-Net is among the strongest-performing architectures, while the ViT becomes increasingly competitive as the training-data budget increases.
Thus, model selection based solely on either synthetic benchmark would favor the CNN, whereas evaluation on the experimental dataset considered here favors different model configurations.

Normalization further modifies synthetic-to-experimental transfer.
The CNN is comparatively insensitive to normalization on the synthetic test sets, but this robustness does not translate into the highest experimental performance.
In contrast, the MDN benefits substantially from z-score normalization: although its synthetic performance remains below that of the CNN, the z-score-normalized MDN becomes one of the strongest-performing models on experimental data.
The MLP likewise benefits from z-score normalization, whereas the U-Net remains comparatively robust to the choice of normalization.
The ViT improves consistently as the training data budget increases under both normalization schemes.
These results indicate that normalization affects not only performance within the synthetic domain but also the extent to which a given architecture transfers from synthetic training to experimental measurements.

The dependence on the training-data budget provides further evidence that synthetic performance alone does not fully characterize experimental behavior.
The CNN improves consistently on both synthetic test sets as the training-data budget increases, whereas its experimental performance varies comparatively little and is not monotonic over the same range.
Conversely, the U-Net and z-score-normalized MDN achieve strong experimental performance even at relatively small training data budgets.
The ViT shows a clearer correspondence between domains, with performance improving as additional training data are introduced on both synthetic and experimental test sets.
Consequently, an increase in synthetic test performance with additional training data does not necessarily translate into a comparable improvement on experimental measurements.

Taken together, these results show that performance within the synthetic domain---even when evaluated on entirely unseen simulated devices---is not sufficient to identify the architecture that performs best on experimental CSDs.
The close agreement between the two synthetic evaluations, contrasted with their substantially weaker correspondence with experimental performance, indicates that the change in model ranking cannot be explained simply by generalization to previously unseen simulated devices.
Instead, the remaining synthetic-to-experimental domain shift plays a central role in determining which architectures and preprocessing choices transfer most effectively to experimental data.

\begin{figure}[t]
  \includegraphics[width=\linewidth]{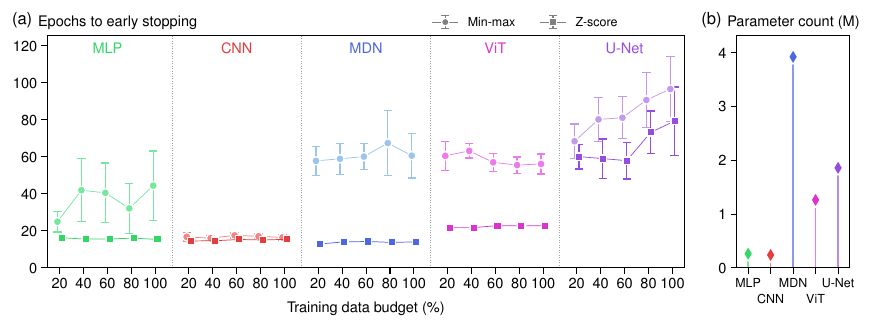}
  \caption{
  (a) Number of training epochs before early stopping based on the validation KL-divergence loss for the MLP, CNN, MDN, ViT, and U-Net as a function of the synthetic training-data budget.
  Circles and squares denote min-max and z-score normalization, respectively; colors identify the model architectures and match those used in figure~\ref{fig:models}.
  Points show the mean across the \num{10} trained models for each condition, and error bars indicate one standard deviation across the \num{10} model-level values, providing a measure of model-to-model variability in training behavior.
  (b) Number of trainable parameters for each architecture, reported in millions (M).
  }
  \label{fig:comp_cost}
\end{figure}

\subsection{Training behavior and performance-size trade-offs}
\label{ssec:computational_efficiency}
Predictive performance alone does not capture the practical differences among the model architectures. 
We therefore also compare their training behavior and model complexity.
Figure~\ref{fig:comp_cost}(a) shows the number of epochs completed before early stopping for each architecture, normalization scheme, and training-data budget. 
The effect of normalization on the required number of training epochs is architecture-dependent. 
For the MLP, MDN, ViT, and U-Net, z-score normalization consistently leads to earlier stopping than min-max normalization.
This difference is particularly pronounced for the MDN and ViT, for which the z-score-normalized models terminate after approximately \numrange{10}{25} epochs across all training-data budgets, compared with roughly \numrange{55}{70} epochs for the corresponding min-max-normalized models. 
The U-Net requires the largest number of training epochs overall, although z-score normalization again reduces the number of epochs relative to min-max normalization.
Differences in stopping epoch therefore reflect architecture- and normalization-dependent convergence under the common KL-based training criterion and should not be interpreted as the epoch at which each model would independently optimize the MSE evaluation metric.

The CNN behaves differently from the other architectures.
Its training terminates after approximately \numrange{15}{20} epochs for both normalization schemes, with comparatively little dependence on either the normalization scheme or the training-data budget.
More generally, the number of epochs required before early stopping varies only weakly with the amount of training data for most model-normalization combinations.
The larger variation observed for the min-max-normalized MLP and U-Net is also accompanied by greater model-to-model variability, whereas the z-score-normalized training runs are comparatively consistent across data budgets.
Because the computational cost of an epoch differs across architectures and training-data budgets, the number of epochs before early stopping should be interpreted as a measure of training behavior rather than a direct measure of total computational cost.

Model complexity also varies substantially among the architectures, see figure~\ref{fig:comp_cost}(b).
The CNN and MLP contain the fewest trainable parameters, with approximately \num{0.24} million and \num{0.26} million parameters, respectively, whereas the MDN is the largest model, with approximately \num{3.9} million.
The ViT and U-Net are intermediate in size, with approximately \num{1.3} million and \num{1.9} million trainable parameters, respectively.
Together with the predictive-performance results in figure~\ref{fig:perf}, these differences motivate considering predictive performance and model complexity jointly when selecting an architecture for experimental deployment.

\begin{figure}[t]
  \includegraphics[width=\linewidth]{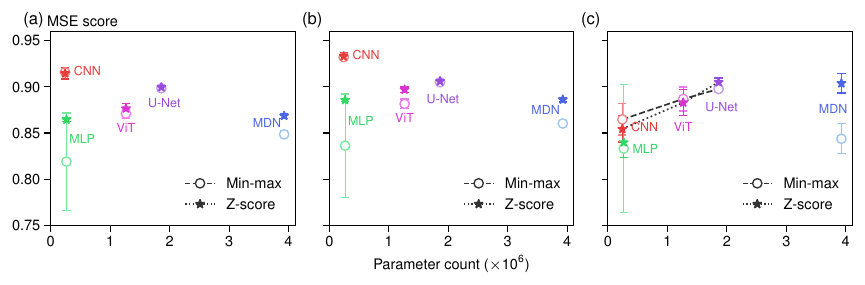}
  \caption{
  Predictive performance as a function of model complexity for
  (a) the held-out synthetic test set,
  (b) the device-held-out synthetic test set, and
  (c) the experimental test set.
  Mean MSE score, $S_{\rm MSE}$, is shown as a function of the number of trainable parameters for models trained using the full synthetic training-data budget.
  Circles and stars show the mean across the \num{10} trained models for each architecture using min-max and z-score normalization, respectively, with error bars indicating one standard deviation.
  Dashed and dotted lines connect the Pareto-optimal architectures for min-max and z-score normalization, respectively, when the corresponding frontier contains more than one model.
  The performance values used in the Pareto analysis follow the same evaluation procedure as in figure~\ref{fig:perf}: synthetic test-set scores are importance-weighted to match the experimental dominant-state distribution, while experimental scores are evaluated without reweighting.
}
  \label{fig:pareto}
\end{figure}

Considering predictive performance together with model size further illustrates the difference between synthetic and experimental model selection.
Figure~\ref{fig:pareto} compares $S_{\rm MSE}$ with the number of trainable parameters for models trained using the full synthetic training-data budget.
On both the held-out and device-held-out synthetic test sets, shown in figure~\ref{fig:pareto}(a,b), the CNN provides the strongest performance-model-size trade-off for both normalization schemes.
It achieves the highest mean score while also having the fewest trainable parameters among the architectures considered.
Consequently, all remaining architectures are dominated in the formal Pareto sense on both synthetic evaluation sets.

The picture changes substantially for experimental data.
Rather than identifying a single preferred architecture, the experimental benchmark produces a sequence of performance-complexity trade-offs; see figure~\ref{fig:pareto}(c).
For min-max-normalized inputs, the CNN, ViT, and U-Net lie on the Pareto frontier, with progressively higher experimental performance as model size increases.
For z-score normalization, the same progression extends to the MDN, which achieves the highest mean experimental score but requires substantially more parameters than the U-Net.
The gain in mean score between the U-Net and MDN is small, however, relative to the difference in model size and the observed model-to-model variability.
Thus, whereas the CNN provides the most favorable performance-size trade-off among the configurations considered on both synthetic benchmarks, evaluation on experimental data reveals a broader set of trade-offs among the architectures.

Because the Pareto frontiers in figure~\ref{fig:pareto} are defined using the mean performance across the \num{10} trained models, we additionally assess the robustness of frontier membership to model-to-model variability using bootstrap resampling.
On both synthetic test sets, the CNN remains Pareto optimal in all bootstrap realizations for both normalization schemes, while none of the other architectures enter the frontier.
For the experimental data, the CNN, ViT, and U-Net exhibit highly robust frontier membership, with $P_{\rm F}\geq 0.998$ for all normalization conditions in which they lie on the mean-score frontier.
In contrast, although the z-score-normalized MDN extends the mean-score frontier beyond the U-Net, it remains Pareto optimal in only approximately \qty{36}{\percent} of bootstrap realizations.
Thus, the bootstrap analysis confirms the robust preference for the CNN on the synthetic benchmarks and the broader performance-complexity trade-off observed experimentally, while indicating that the additional frontier extension provided by the MDN is comparatively sensitive to model-to-model variability.

\section{Conclusion and outlook}
\label{sec:conclusion}
In this work, we benchmark five machine-learning architectures for state recognition in double-QD CSDs, examining how model architecture, input normalization, and the amount of synthetic training data affect predictive performance and transfer to experimental measurements.
By training all models using the same simulated data and evaluating them on common synthetic and experimental test sets, we distinguish performance within the simulation domain from the ability of the resulting models to generalize to experimentally acquired CSDs.
Our results show that these two objectives are not equivalent: synthetic benchmark performance alone does not identify the model that transfers best to experimental data.

The differences among architectures are apparent already in their data efficiency.
The U-Net reaches nearly its maximum synthetic performance using only a fraction of the available training data, whereas the MLP, MDN, and ViT continue to benefit from additional synthetic examples over the range considered.
The CNN achieves the strongest performance on both synthetic benchmarks while remaining comparatively insensitive to the training data budget and normalization.
This advantage does not carry directly to experimental data, however.
The U-Net transfers particularly well despite ranking below the CNN on synthetic data, while the z-score-normalized MDN achieves unexpectedly strong experimental performance.
The ViT also becomes increasingly competitive as the amount of training data increases.
Thus, both architecture and preprocessing influence not only performance within the synthetic domain, but also the extent to which that performance transfers to experiments.

Considering model complexity together with predictive performance further emphasizes this difference.
On both synthetic test sets, the CNN provides a particularly favorable performance-complexity trade-off, achieving the highest mean score while also having the fewest trainable parameters among the architectures considered.
Experimental evaluation instead produces a broader Pareto frontier.
Increasing model complexity from the CNN through the ViT and U-Net yields progressively higher experimental performance, while the z-score-normalized MDN provides an additional, albeit comparatively small, improvement at substantially greater model complexity.
Bootstrap analysis confirms that the preference for the CNN is robust on both synthetic test sets, whereas the additional experimental trade-off provided by the MDN is more sensitive to model-to-model variability.
Importantly, evaluation on device-held-out synthetic data closely reproduces the conclusions obtained from the standard held-out synthetic benchmark.
The change in model ranking observed for experimental data therefore cannot be explained simply by generalization to previously unseen simulated devices.
Rather, it reflects the remaining synthetic-to-experimental domain shift.

Taken together, these results indicate that no single architecture is optimal across synthetic performance, experimental transfer, data efficiency, and model complexity.
The appropriate model instead depends on the intended use.
The CNN provides a compact and robust baseline, particularly when model size and strong performance within the simulated domain are priorities.
The U-Net offers strong experimental transfer and high data efficiency at greater model complexity, while the MDN demonstrates that an architecture that is not competitive on the synthetic benchmark can nevertheless transfer particularly well when paired with an appropriate normalization scheme.
The ViT provides another competitive experimental option, although it benefits more strongly from increasing the synthetic training-data budget.
More broadly, these results show that selecting a model solely according to performance on simulated test data can lead to a different choice than selecting one for deployment on experimental measurements.

Several extensions of this benchmark are natural.
Expanding the evaluation to additional experimental devices, material platforms, and larger QD arrays will be important for determining how broadly the observed model rankings and synthetic-to-experimental transfer behavior persist.
Increasing the diversity of simulated devices and noise models may also reduce the remaining domain gap and help identify synthetic evaluation protocols that better predict experimental performance.
Domain adaptation and transfer-learning approaches, including fine-tuning on small experimental datasets and incorporating physics-informed constraints, provide complementary strategies for improving transfer when sufficiently representative simulation is difficult to achieve.
Finally, extending the benchmark to include uncertainty quantification and interpretable model outputs would provide additional criteria for evaluating models intended for autonomous operation of quantum devices.
Together, these directions can move model benchmarking beyond identifying the best-performing architecture on a simulated test set toward establishing which models are sufficiently reliable, efficient, and transferable for practical automated control.

\ack{
This research was sponsored in part by the U.S. Army Research Office (ARO) under Award Nos. W911NF-23-1-0258 and W911NF-24-2-0043.
The views and conclusions contained in this paper are those of the authors and should not be interpreted as representing the official policies, either expressed or implied, of the ARO or the U.S. Government. 
The U.S. Government is authorized to reproduce and distribute reprints for Government purposes, notwithstanding any copyright noted herein. 
Certain equipment, instruments, software, or materials are identified in this paper in order to specify the experimental procedure adequately.  
Such identification is not intended to imply recommendation or endorsement of any product or service by the National Institute of Standards and Technology, nor is it intended to imply that the materials or equipment identified are necessarily the best available for the purpose.
}


\end{document}